\documentclass[11pt,a4paper]{article}
\usepackage[hyperref]{emnlp2020}
\usepackage{times}
\usepackage{latexsym}
\usepackage{graphicx}
\usepackage{siunitx}
\usepackage{tikz-dependency}

\usepackage{microtype}

\aclfinalcopy 

\title{Measuring Information Propagation in Literary Social Networks}

\author{

   \bf Matthew Sims \\
   School of Information \\
   UC Berkeley \\
   \texttt{msims@berkeley.edu} \\\And
  
   \bf David Bamman \\
   School of Information \\
   UC Berkeley \\
   \texttt{dbamman@berkeley.edu} \\
}

\date{}

\begin{document}
\maketitle
\begin{abstract}
We present the task of modeling information propagation in literature, in which we seek to identify pieces of information passing from character $A$ to character $B$ to character $C$, only given a description of their activity in text. We describe a new pipeline for measuring information propagation in this domain and publish a new dataset for speaker attribution, enabling the evaluation of an important component of this pipeline on a wider range of literary texts than previously studied.  Using this pipeline, we analyze the dynamics of information propagation in over 5,000 works of English fiction, finding that information flows through characters that fill structural holes connecting different communities, and that characters who are women are depicted as filling this role much more frequently than characters who are men.

\end{abstract}

\section{Introduction}

With the rise of sociological approaches to narrative, work in  literary criticism has increasingly turned to the ways in which authors depict social networks in their texts. This includes critical attention to both \textit{network topologies}, such as understanding characters and their structural relationships with others\ \citep{levine2009}, and \textit{information flow}, such as theorizing the representation of disease and gossip\ \citep{levine2009, margolis2012,spacks85}.
Much computational work in NLP has arisen to support the former line of research, including extracting social networks from text\ \citep{elson:2010:esn:1858681.1858696}, predicting familial relationships\ \citep{DBLP:journals/corr/MakazhanovBK14}, and modeling the interactions between characters\ \citep{Iyyer:Guha:Chaturvedi:Boyd-Graber:Daume-III-2016,Chaturvedi:Iyyer:Daume-III-2016}. This in turn has driven work in the digital humanities examining the structure of literary networks \citep{moretti2011network, algeehewitt2017,piper2017,alexander2019}.

\begin{figure}[!htb]
\includegraphics[scale=.38]{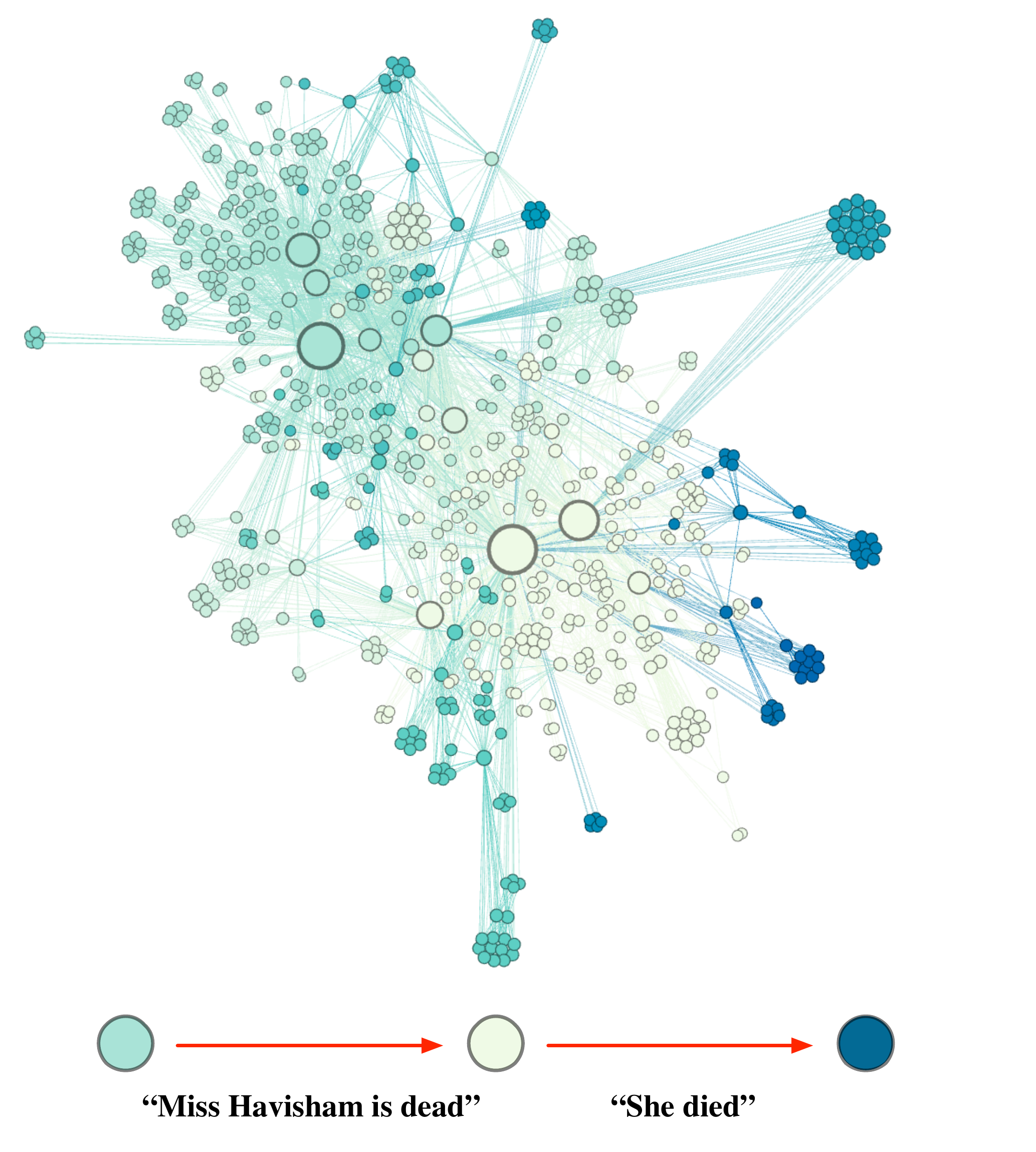}
\caption{\label{network_viz} The character co-occurrence network for \textit{Great Expectations}. Nodes represent characters and edges represent conversational interactions. Below the network, we illustrate an example of information transmission across a character triad. 
}
\end{figure}

At the same time, however, there remains a substantial gap in computational work to support the latter research goal of modeling the flow of information within depicted networks.
Yet understanding how the transmission of information is represented in these imagined worlds has the potential to be of great value to scholars in the humanities, since  the resulting models can serve as a basis for broader insights about the social structures embedded in narratives, the role of characters based on  attributes such as race and gender, and the informational dynamics of gossip\ \cite{spacks82,spacks85,martin2014}.

 In this work, we specifically aim to fill this gap by developing methods to track the flow of information in novels by extracting instances of a message passing from character $A$ to character $B$ to character $C$, only given a depiction of their conversational interactions. We develop a methodology for modeling this mode of propagation in both \emph{explicit} networks (where one character provides information that is explicitly attributed to another character, such as ``Bob told me that Jack escaped''); and in \emph{implicit} networks, where information is repeated by multiple characters without such attribution. While the results of the methods enable a range of potential analyses---for instance, comparative analysis between authors, characters, and dyads---we focus on two illustrative case studies. First, we examine the linchpins of information flow---the characters who are most responsible for the propagation of information---and how they are positioned relative to the overall network topology; and second, we examine the gender dynamics of information propagation and what it tells us about how novelists represent men and women as the means and agents for transmitting facts, gossip, and other details about the social workings of these imagined worlds.

 We make the following contributions with this work:
 
 \begin{enumerate}
\item We present a new NLP pipeline for determining information propagation in literary texts, incorporating a range of different sub-tasks, including coreference resolution, speaker attribution, character network identification, and information extraction. 

\item We present a new dataset for speaker attribution, comprised of 1,765 quotations linked to their speakers in 100 different literary texts, allowing us to evaluate a critical component of this pipeline on a wider range of literary texts than previously studied.  

\item We leverage our pipeline to analyze the dynamics of information propagation in a collection of 5,345 works of English fiction from Project Gutenberg. We find that information flows through characters that fill structural holes connecting different communities, and that characters who are women are depicted as filling this role much more frequently than characters who are men.

\end{enumerate}

\section{Related work}

Much of the computational research into information propagation and diffusion has focused on the domain of social media \citep{DBLP:journals/corr/abs-1201-4145}.
Research in this area includes analyses of information diffusion in blogs \citep{gruhl2004information,leskovec2007patterns}, the spread of news across online networks \citep{leskovec2009meme}, and in particular, the spread of rumor and misinformation \citep{kwon2013prominent,friggeri2014rumor,del2016spreading,vosoughi2018spread}. 

A core aspect of this work that strongly differs from networks in fiction is that the individual components of social media networks (the nodes, edges, and instances of propagation) are often directly observed. In modeling retweet dynamics in Twitter, for instance, nodes are defined as unique users, edges are directly observed friend and follow links defined by the platform, and propagation occurs when one user retweets a message posted by another they are connected to. More closely related to the challenges posed by detecting propagation in fiction is work that may directly observe the node and edge structure of a network, but must infer an act of \emph{propagation}, including work in tracking the diffusion of memes \citep{leskovec2009meme}, text reuse across legislative bills\ \citep{wilkerson2015tracing} and quotations in news\ \citep{niculae2015quotus}.

While information propagation has yet to inform work in narrative (hence the purpose of this study), network structure has increasingly informed literary scholarship.
Following the work of \citet{bourdieu1996rules},
literary scholars have in recent years begun to explore the role that social networks play both in authorial composition \citep{so2013network,mazanec2018networks} and in the narrative representation of ``networked social experience''\ \citep{levine2009}.

Treating literary works \textit{themselves} as networks, however, poses distinct computational challenges. While research into information propagation in social media tends to presume access to explicit networks, the character networks represented in novels are implicit.  To determine these networks, we draw on previous work by\ \citet{elson:2010:esn:1858681.1858696}, who build edges between character nodes through conversational interactions.
Various computational work to extract social networks from literature has built on this research over the past ten years,\footnote{See \citet{labatut} for a review.} including fundamental methods designed to extract networks for other languages like German\ \citep{jannidis2016}, incorporate other categories of nodes such as locations\ \citep{lee-yeung-2012-extracting} and objects\ \citep{Sudhahar2013}, and analyze the structure of networks to test specific hypotheses\ \citep{elson:2010:esn:1858681.1858696,agarwal-etal-2012-social,coll-ardanuy-sporleder-2014-structure,piper2017}. Our work builds on this tradition by introducing methods to reason about the phenomenon of propagation in fiction based on these constructed networks.

\section{Methods}

Our goal in this work is to investigate the behavior of information propagation in literary texts. In order to identify acts of propagation in this context, we need to determine the underlying network structure of a novel, including the nodes (by inferring characters) and the edges (by inferring some interaction between them).
We describe first our pipeline for doing so, which involves identifying a set of unique characters from their mention in a text using coreference resolution (\S\ref{s:coref}), attributing dialogue to those characters (\S\ref{s:speaker}), building a social network of speakers and listeners from that data (\S\ref{s:buildnetworks}), and operationalizing a measure of ``information'' that we can treat as an atomic unit involved in propagation using slot-based information extraction (\S\ref{s:information}).  
With these constructed networks, we can measure acts of implicit propagation (\S\ref{s:implicitprop}) and explicit propagation (\S\ref{s:explicitprop}) within it.

\subsection{Coreference resolution}\label{s:coref}

Most contemporary systems for coreference resolution are trained on the benchmark OntoNotes dataset\ \citep{hovy_ontonotes:_2006}, which primarily consists of news and conversation; literature is represented there only in the  narrow genre of the Bible. 

In order to use coreference resolution specifically trained on literature,  we use the coreference annotations and trained model described in \citet{bamman2019annotated}.  This model is a neural coreference system inspired by \citet{lee-etal-2017-end}, augmented with BERT contextual representations\ \citep{devlin-etal-2019-bert}, and trained on 210,532 tokens in LitBank, comprising 100 different works of English-language fiction. \citet{bamman2019annotated} report its cross-validated average F-score on LitBank to be 68.1, notably higher than the performance for a model trained on OntoNotes (which has an average F1 score of 62.9).

\begin{table*}[h!]
\centering
\begin{tabular}{|l|c|c|c|c|c|} \hline
&$B^3$&MUC&CEAF$_{\phi_4}$&Average&$\Delta$\ \\ \hline
Predicted coreference&68.0&84.9&61.0&71.3&-- \\ \hline
$\;$--Trigram matching&67.7&84.5&60.1&70.8&-0.5 \\ \hline
$\;$--Dependency parses&66.5&83.4&56.8&68.9&-2.4 \\ \hline
$\;$--Singleton mention detection&67.0&85.9&59.6&70.8&-0.5 \\ \hline
$\;$--Paragraph final mention linking&68.0&84.9&61.0&71.3&0.0 \\ \hline
$\;$--Vocatives&68.9&85.9&62.1&72.3&+1.0 \\ \hline
$\;$--Conversational pattern&66.8&85.0&58.9&70.3&-1.0 \\ \hline\hline
Oracle coreference &80.2&89.7&74.7&81.5&+10.2 \\ \hline

\end{tabular}
\caption{\label{speaker} Metrics for cluster overlap between the gold set of clusters $\mathcal{G}$ and predicted set of clusters $\mathcal{C}$.  Each cluster is defined as the set of quotations spoken by the same speaker. We also present the upper bound of carrying out speaker attribution using gold coreference labels (oracle coreference), which suggests that there is much to be gained in improving quotation attribution not only by improving coreference, but independently of it as well.}
\end{table*}

\subsection{Speaker attribution}\label{s:speaker}

\paragraph{Data.}

Previous work in literary speaker attribution has focused on a relatively small set of novels. Both \citet{he-barbosa-kondrak:2013:ACL2013} and \ \citet{muzny2017two} annotate Austen's \emph{Pride and Prejudice} and \emph{Emma} as well as Chekhov's \emph{The Steppe}. Similarly, the Columbia Quoted Speech Corpus includes six texts by Austen, Dickens, Flaubert, Doyle and Chekhov. While these datasets have been able to drive much work in the development of models for speaker attribution, they represent a comparatively narrow slice of how dialogue is depicted in literature.

In order to evaluate the robustness of models across a diverse range of novels and authors, we annotate all 100 texts in LitBank\ \citep{literaryentities} with the  boundaries for all true quotations and link each to the {entity} who spoke it. Here we are able to draw on the coreference annotations present in LitBank, which already link each mention to a unique entity. All annotations were carried out using the BRAT annotation interface\ \citep{Stenetorp:2012:BWT:2380921.2380942} by four annotators after a period of initial training, prompted to identify all quotations and attribute each one to the speaker who uttered it.  Given the high agreement rate observed by\ \citet{muzny2017two} ($\kappa$ of 0.97 for quote-speaker labels), each quotation is attributed by a single annotator. To check consistency, we double-annotate a sample of 10 texts (10\% of the entire collection) at the end of the annotation process and find a similarly high inter-annotator agreement rate (Cohen's $\kappa$ of 0.962). In total, 1765 quotations were annotated across all 100 works of fiction. This data is freely available under a Creative Commons ShareAlike 4.0 license at \url{https://github.com/dbamman/litbank}.

\paragraph{Quotation identification.}
For the task of quotation identification, we use the method implemented in BookNLP\ \citep{bammanliterary2014}, which uses simple regular expressions (text contained between an opening quote and a closing quote).  On our gold annotations, this method results in an F1 score of 90.8 for quotation identification (87.1 precision and 95.0 recall). 
False positive failure cases of strings wrapped in quotation marks that do not constitute dialogue include various typographical uses of quotation for signifying other phenomena, including scare quotes for emphatic use (to introduce jargon, neologisms, or irony), titles of works of art, the mention of a term (as distinct from its use), and written use (see \citet{quote2011} for a survey). False negatives primarily arise due to regex matching errors (such as a stray quotation mark that results in an inversion of the subsequent speech and narration), or texts that do not delimit speech with quotation marks at all (such as Joyce's \emph{Ulysses}, which introduces direct speech with dashes).

\paragraph{Attribution.}

For speaker attribution, we reimplement the deterministic method of \citet{muzny2017two} using the full coreference information predicted above. \citet{muzny2017two} describes a series of deterministic sieves for the two tasks of a.) mapping quotes to the nearest speaker mention and b.) linking identified speaker mentions to character entities.  The Quote$\rightarrow$Mention phase includes sieves such as high-precision regular expressions  for predefined Quote/Mention/Verb patterns (e.g., [``\ldots,'']$_{QUOTE}$ [said]$_{VERB}$ [Jane]$_{MENTION}$), originally defined in \citet{Elson:2010:AAQ:2898607.2898769}; dependency structure information (identifying mentions that hold an \textsc{nsubj} relation to a verb of communicating); and vocatives in the previous quotation. Quotations unattributed after running all sieves are assigned the majority speaker in the context.

To separate out the task of quotation identification from quotation attribution, we evaluate quotation attribution with gold quotation boundaries. While previous work on quotation attribution in literary texts, including \citet{muzny2017two} and \citet{he-barbosa-kondrak:2013:ACL2013}, evaluate system performance using classification accuracy and precision/recall (where each quotation in a test book is judged to be assigned to the correct true speaker from a predefined gold character list), we do not presume access to a gold character list during prediction time. Like \citet{almeida-etal-2014-joint}, we evaluate performance using a measure of cluster overlap (here, the suite of metrics used in evaluating coreference resolution), where each cluster is defined by the set of quotations spoken by the same speaker.

As Table \ref{speaker} illustrates, our reimplementation of \citet{muzny2017two} for the task of speaker attribution yields an average F1 score of 71.3 across the three cluster metrics when evaluated on all 100 books in our newly annotated data.  As we ablate different aspects of the \citet{muzny2017two} model, performance generally degrades, attesting to the value of individual sieves.

\subsection{Identifying character networks}\label{s:buildnetworks}

Similar to previous approaches for determining character networks in literary fiction  \citep{Elson:2010:AAQ:2898607.2898769,moretti2011network}, we use conversation as the basis for determining the edges in our graph. However, rather than trying to identify specific speaker-listener interactions, we instead extract dialogue blocks, drawing an edge between all characters mentioned outside of a quotation in a given block. Edges are weighted by the total number of dialogue blocks in which a given pair of characters are found to be co-present. We use a simple heuristic to identify these conversation blocks: if three or more contiguous sentences do not contain any quoted dialogue, we treat this as the termination of the block. The resulting graph serves as the basis for identifying information propagation in a given novel, as detailed in the following subsections.

\subsection{Defining information}
\label{s:information}

Whereas large-scale corpora such as social media data sets provide networks in which fuzzy matching of information may be appropriate and in which information repetition is often substantial\ \citep{leskovec2009meme}, in the context of novels such methods are unlikely to have sufficient precision. As a result, we select an information approach which allows us to maximize precision at the cost of  potentially missing some instances of propagation. Our approach entails identifying quoted speech that references at least one character. One way to define this type of speech would be to simply describe it as \textit{gossip}, though we feel that this is an overly narrow term given the general nature of our approach. Specifically, we select propositional tuples of the form (subject, verb, object), such that the subject holds an \texttt{nsubj} dependency relation to the verb and the object holds an \texttt{obj} relation (using the terminology of the Universal Dependencies\ \citep{nivre2016universal}); the subject and object may be filled by a character entity, a non-character nominal phrase, or a null token if neither is present. We ignore any tuples which contain \textit{I}, \textit{you}, or \textit{we} (along with their variants) in either the subject or object slots, since they have comparatively higher errors in coreference. For the verb slot, we always select the lemma form of the proposition's head verb.  Character entities in a proposition are identified by their unique character IDs established through coreference resolution (and not by the surface form of their mention). 

Consider the following hypothetical example:
\\[1pt]
\begin{center}
\begin{dependency}[theme = default,
label style={font=\bfseries,thick}
]
   \begin{deptext}[column sep=-.3em, row sep=.0ex]

Bob \& punched \& Tom \& and\& he \& left \\
   \end{deptext}
   \depedge{2}{1}{nsubj}
   \depedge{2}{3}{obj}
    \depedge{6}{5}{nsubj}
\end{dependency}
\end{center}

Given the operation of coreference resolution  mapping ``Bob'' and ``he'' to the entity \texttt{Bob-id1} and ``Tom'' to \texttt{Tom-id2}, the extracted tuples for this sentence would be: [\texttt{Bob-id1}, punch, \texttt{Tom-id2}] and [\texttt{Bob-id1}, leave, $\emptyset$]. We extract all propositional tuples using a set of rules applied to the dependency parse of a given sentence. Although reductive to some degree, defining and extracting information in this way allows us to avoid informational noise and only select consistent propositional units. 

To further reduce potential informational noise, we also only select tuples containing words that are likely to have some intrinsic interest to the plot and which have a relatively fixed meaning. After analyzing the 100 words that occur most frequently across all the tuples extracted from our corpus, we select tuples containing terms associated with the following four categories: \textit{amorous}, \textit{hostile}, \textit{juridical}, and \textit{vital}. For each category, we include the following words along with any synonyms that are also present in the top 100 tuple words: \textit{amorous} (love, marriage), \textit{hostile} (hurt, hit, shoot, kill), \textit{juridical} (arrest, escape, innocent, guilty), and \textit{vital} (alive, sick, dead).  Since the Gutenberg corpus primarily contains nineteenth-century novels, these topics reflect many of the key events that these works of fiction tend to focalize.

\subsection{Defining implicit propagation} 
\label{s:implicitprop}

We identify instances of implicit information propagation simply by determining whether a propositional tuple passes between a minimum of three characters. In other words, we look for an informational triad of the form character $A$ $\rightarrow$ character $B$ $\rightarrow$ character $C$, such that character $A$ and character $B$ are co-present when character $A$ voices the initial instance of the proposition (but character $C$ is not), and character $B$ and character $C$ are co-present when character $B$ repeats the proposition during a different conversation block.

\subsection{Defining explicit propagation}
\label{s:explicitprop}

Along with implicit instances of information propagation, we note that novels often contain explicit propagation as well. We define explicit propagation as occurring when a character reports what another character said to a third character. In other words, we simply search for  variations of the pattern ``\texttt{[character-id]} said" in the context of quoted speech. Specifically, the variations considered include synonyms for ``say" along with any arguments or modifiers that are relevant to introducing reported speech (e.g., ``declared," ``told me," ``mentioned that," ``claimed to," etc.).

The benefit to capturing instances of explicit propagation is that such instances can be extracted with very high precision regardless of the informational topic being discussed. Consequently, unlike for implicit propagation, we make no constraints on the nature of the information itself (in contrast to the four topics defined above). After identifying instances of explicit propagation, we incorporate coreference resolution and speaker attribution to determine the specific characters of a given propagating triad.  Section \ref{explicit_analysis} discusses how the resulting data from this approach can be used to analyze the role that gender plays in the depiction of information propagation within novels.

\section{Experiments}
\label{experiments}

Given instances of information propagation extracted from novels, we seek to understand the structural roles of the literary network and its topography that contribute to information passing between dyads.  In particular, we seek to disentangle two possible alternatives:

\begin{itemize}
    \item[H1.] Information propagates through  bridges who pass information between otherwise disconnected communities. 
    \item[H2.] Information propagates among densely connected strong ties (such as between members of the same family who interact frequently).
\end{itemize}

These alternatives correspond to different functions of gossip in literature, as theorized by \citet{spacks85}: while gossip primarily involves the ``deliberate circulation of information,'' it also functions to reinforce existing relationships among strong ties (a point taken up in real-world social networks in \citet{foster2004}).
We operationalize this distinction for understanding the dynamics of {implicit propagation} by describing information-propagating characters and non-information-propagating characters through six network measures that capture their topological properties within the network:

\begin{enumerate}

\item \textbf{Closeness centrality}: the average inverse distance between a given node and all other nodes in a graph. 
\item \textbf{Betweenness centrality}: the fraction of shortest paths that pass through a node, summed over all node pairs.

\item \textbf{Average neighbor degree}: the average degree of the nodes in a given node's neighborhood.

\item \textbf{Effective size}: the measure of the non-redundancy between a node and its contacts---specifically how connected a node's contacts would be in its absence (i.e., the resulting structural hole).

\item \textbf{Efficiency}: the effective size of a node divided by its degree.

\item \textbf{Triangle count}: the number of triangles for which a node serves as a vertex, where a triangle is defined as a set of three nodes that are directly connected to each other.

\end{enumerate}

We use the above measures to describe all nodes that either function as the $B$ node in an $A\rightarrow B\rightarrow C$ information triad, or could function as such a node. More specifically, whenever we observe an instance of propagation $A\rightarrow B\rightarrow C$, in which at least one separate character $B'$ was co-present with $B$ when hearing $A$'s information (but did not propagate it further), we select a pair comprised of $B$ as a propagating node and $B'$ as a non-propagating node (sampling $B'$ from the set of co-present characters if more than one was present). In cases in which no non-propagating character was co-present, we instead sample a $B'$ from the set of propagating instances that had multiple co-present characters.   When sampling the non-propagating $B'$ nodes, we only select characters that have been observed to speak at least once in the text based on our speaker attribution model (we hypothesize that selecting these characters is a better way to judge the efficacy of a propagation model, since they at least vocalize some information in the narrative, and hence are more likely to resemble propagating nodes in terms of their narrative functions).

\subsection{Results}

In order to test implicit information propagation in literature, we run tuple extraction on 5,345 works of fiction from the Project Gutenberg corpus. We find that roughly 3,600 of these books contain at least one instance of a repeated tuple containing a word from our four topics of interest (indicating the possibility of propagation based on our criteria). We proceed to run the rest of our pipeline on this subset of books.\footnote{We process all books on a high-performance computing cluster using 24-core Intel Xeon Haswell processors and 64 GB of RAM; the average runtime of this pipeline on one book on this platform is five minutes and two seconds.} In total, we find that 35\% of these works contain at least one instance of implicit information propagation.

To distinguish between the two hypotheses outlined above, we scale all the features of the  data between 0 and 1 and train a non-regularized logistic regression model to distinguish between information propagating $B$ nodes and non-propagating $B'$ nodes. We run the model on 1,730 $B$ nodes and 1,730 $B'$ nodes.  The results are shown in Table \ref{graph_measures} and discussed in more detail in the next section.

\begin{table}[h!]
\centering
\begin{tabular}{|c|c|} \hline
\textbf{Graph Measure} & \textbf{Coefficient}  \\ \hline
Efficiency &$\phantom{-}3.0^\ast$   \\ \hline
Effective size &$\phantom{-}2.7\phantom{^\ast}$   \\ \hline
Betweenness centrality &$\phantom{-}0.5\phantom{^\ast}$  \\ \hline
Closeness centrality &$\phantom{-}0.1\phantom{^\ast}$  \\ \hline

Triangles &$-0.4\phantom{^\ast}$\\ \hline

Average neighbor degree &$-4.9^\ast$    \\ \hline
\end{tabular}
\caption{\label{graph_measures} Logistic regression model coefficient values. Stronger positive values are indicative of information-propagating nodes; stronger negative values are indicative of non-propagating nodes. $^\ast$ denotes $p < 0.01$.}
\end{table}

To ensure that our results are not simply caused by aspects of each  network's general topology (irrespective of the unique qualities of propagating $B$ nodes) we also run a degree-preserving randomization experiment\ \citep{10.1007/978-3-642-21286-4_10} as a more stringent means for testing significance. For each network containing a propagating node, we generate 10 expected degree graphs and use them to calculate network measures for the corresponding propagating $B$ and non-propagating $B'$ nodes in the original network, producing a set of 10 randomized measures for each of the 1,730 original nodes in each class. We then randomly sample a single measure from each of these sets, yielding 1,730 randomized node measures for both classes, and re-run our logistic regression model on that resample. We repeat this process 10,000 times to generate an expected null distribution for each coefficient and assess the frequency with which a null coefficient value was observed to be as extreme as the value we observe under the true network---analogous to a p-value in a bootstrap hypothesis test\ \citep{efron1982jackknife,berg2012empirical,dror-etal-2018-hitchhikers}.

For the two node measures found to be significant under our original model, efficiency has a p-value of 0.08 (8\% of 10,000 random trials observe a statistic as extreme as $3.0$), no longer rising to the level of significance at $\alpha=0.01$, while average neighbor degree has a p-value of 0 (no random trial sees as a statistic as extreme as $-4.9$), providing further evidence of its significance as a feature for discriminating information-propagating nodes.

\section{Analysis}

\subsection{Implicit propagation and weak ties}

As Table \ref{graph_measures} shows, average neighbor degree and efficiency are both found to be significant at a threshold of $\alpha=0.01$, while average neighbor degree is confirmed to be significant under a degree-preserving randomization experiment. These results support the first of our two postulated hypotheses (introduced in \S \ref{experiments}): information in novels propagates through characters that serve as bridges between otherwise disconnected communities.  

Average neighbor degree has the largest coefficient (by absolute value) and is negatively correlated with propagation. 
High values of average neighbor degree denote communities that are already well-connected (both to each other and to the rest of the network).  
In such a information-rich neighborhood, instances of propagation would be of less value or necessity, and hence would be less likely to be observed. 

Support for the first hypothesis is further confirmed by the strong positive coefficient for efficiency. Like effective size, efficiency is a means of determining the extent to which a structural hole would occur if a specific node were removed from the network. Whereas effective size indicates the possibility of such a structural hole in general, efficiency measures how much each one of a node's connections  on average contribute to linking otherwise disconnected neighborhoods. Thus high efficiency suggests that a node is not only serving as a useful bridge between other nodes, but that it is doing so productively relative to its total number of connections.

In a sense, these results suggest that we are observing a version of the weak tie theory first proposed by \citet{granovetter1973strength}. By virtue of the fact that a character's connections are not themselves closely connected, that character can in turn serve an essential informational role for the community.

\subsection{Explicit propagation and gender}

While our methods for extracting implicit propagation for amorous, hostile, juridical and vital events identified 1,730 instances in 5,345 novels, our method for identifying explicit propagation yields far more---93,948 instances of propagation involving 258,619 triads (since there may be multiple listeners for a single instance). Although the analysis carried out on implicit propagation is not possible for the explicit case (since there is no way to identify co-present $B'$ nodes when the initial instance of a proposition remains unobserved in the text), the size of the explicit results are conducive to other analyses. Specifically, we consider here the role that gender plays in the depiction of propagation.  As \citet{spacks85} points out, women are often stereotyped (both within the real world and in representations in literature) as more likely to engage in gossip; from a networked perspective, they are also often cast as intermediaries between men, ``serving as points through which to triangulate male-to-male desire or power'' \citep{Selisker2015}.
Analyzing gender (and other demographic attributes) in the context of information propagation enables scholars to consider how authors construct the informational ecology of their novels given the functional roles played by different characters. 

\label{explicit_analysis}

To measure the role that gender plays in how authors represent information propagation in novels, we calculate the relative proportion of different gender configurations for propagating triads compared to all triads present across our entire data set (we determine the gender of a character by counting up all the male and female nouns and pronouns in that character's coreference chain). This allows us to answer the question: given the overall structural opportunity to transmit information, how often does transmission actually occur based on gender?

\begin{figure}[!htb]
\begin{centering}
\includegraphics[scale=.65]{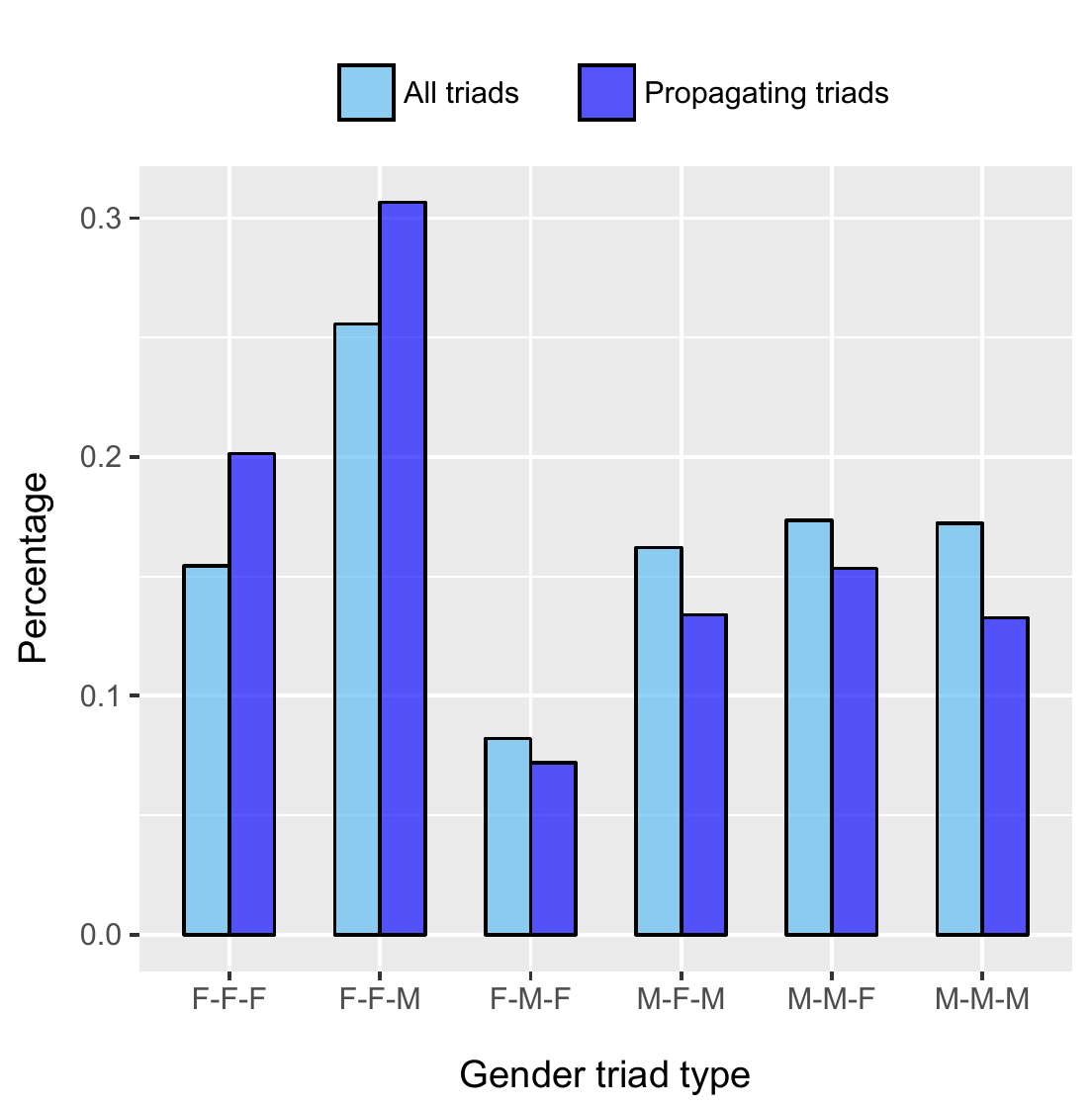}
\caption{\label{gender} Comparison of the relative proportions of triad variations based on gender. All triads (light blue, $n=$ 158,250,238) represent every observed triad across 5,345 books. Propagating triads (dark blue, $n=$ 258,619) indicate only those triads observed to explicitly propagate information.  The widest 95\% confidence interval across all proportions is $\pm 0.0018$, so that all differences within a gender triad type are significant.}
\end{centering}
\end{figure}

Figure \ref{gender} illustrates the proportion of each gender configuration compared to the total; for instance, while 15.4\% of all character triads are comprised of three women (\texttt{F-F-F}), 20.1\% of triads involved in information propagation involve three women. Overall, we find that not only are female characters more likely to serve as propagators than male characters in this dataset, but that female characters fill this role more frequently than one would expect given the proportion of female connector nodes across all triads.  The proportion of propagation when the middle node is male, conversely, is lower than the expected value for every configuration. In other words, authors represent women as information propagators comparatively more frequently than men relative to their overall expectation. 

Although literary criticism tends to envision the role of women in novels as being intermediaries between men\ \citep{woolf1929,Sedgwick1985, schantz2008,Selisker2015}, our analysis of information propagation tells a slightly different story. While women do indeed appear to serve as intermediaries/connectors more frequently than men do, women  propagate information \emph{between} men much more rarely than they do in other configurations (i.e., \texttt{F-F-F}, \texttt{F-F-M}). Though women may of course still connect men in these narratives, they do not appear to do so by notably passing on information.  We leave determining the broader significance of this result to future work.

\section{Conclusion}

We introduce the task of identifying information propagation in literary social networks, designing an NLP pipeline for extracting both \emph{implicit} and \emph{explicit} propagation.  This work offers a new perspective on the analysis of social networks in literary texts by considering the dynamics of how information flows through them---both as a result of the structural topology of the network (characters who successfully propagate are information bridges between communities), and as a result of the specific characteristics of each node (women are depicted more frequently as successful propagators than men). 

This study, of course, contains limitations: readers of fictional works are only afforded a partial perspective of the world that is represented---namely the interactions an author chooses to describe (and not, for example, the dialogue we might presume takes place ``off-screen"). Considered from a narratological perspective, however, this is a benefit rather than a drawback, since our goal is not to understand the underlying reality of these imagined worlds but rather how authors opt to represent the informational dynamics from which their stories are constructed.  In developing this pipeline to examine how authors depict the transmission of information within narrative texts, we hope to drive a variety of future research in this space, including not only such narratological questions as how ``gossip impels plots''\ \citep{spacks85}, but also questions pertaining to issues of bias in representation, the flow of information, and factuality. Code to support this work can be found at \url{https://github.com/mbwsims/literary-information-propagation}.

\section*{Acknowledgments}

Many thanks to Olivia Lewke, Anya Mansoor and Sam Levin for research support and to Lucy Li, Richard Jean So and the anonymous reviewers for valuable feedback on this work. The research reported in this article was supported by funding from the National Science Foundation (IIS-1942591) and the National Endowment for the Humanities (HAA-256044-17), and by resources provided by NVIDIA.

\bibliographystyle{acl_natbib}
\bibliography{infoprop}

\begin{thebibliography}{55}
\expandafter\ifx\csname natexlab\endcsname\relax\def\natexlab#1{#1}\fi

\bibitem[{Agarwal et~al.(2012)Agarwal, Corvalan, Jensen, and
  Rambow}]{agarwal-etal-2012-social}
Apoorv Agarwal, Augusto Corvalan, Jacob Jensen, and Owen Rambow. 2012.
\newblock Social network analysis of alice in wonderland.
\newblock In \emph{Proceedings of the {NAACL}-{HLT} 2012 Workshop on
  Computational Linguistics for Literature}, pages 88--96, Montr{\'e}al,
  Canada. Association for Computational Linguistics.

\bibitem[{Alexander(2019)}]{alexander2019}
Sam Alexander. 2019.
\newblock Social network analysis and the scale of modernist fiction.
\newblock \emph{Modernism/modernity}, 3.

\bibitem[{Algee-Hewitt(2017)}]{algeehewitt2017}
Mark Algee-Hewitt. 2017.
\newblock Distributed character: Quantitative models of the {E}nglish stage,
  1550--1900.
\newblock \emph{New Literary History}, 48(4).

\bibitem[{Almeida et~al.(2014)Almeida, Almeida, and
  Martins}]{almeida-etal-2014-joint}
Mariana S.~C. Almeida, Miguel~B. Almeida, and Andr{\'e} F.~T. Martins. 2014.
\newblock A joint model for quotation attribution and coreference resolution.
\newblock In \emph{Proceedings of the 14th Conference of the {E}uropean Chapter
  of the Association for Computational Linguistics}, pages 39--48, Gothenburg,
  Sweden. Association for Computational Linguistics.

\bibitem[{Bakshy et~al.(2012)Bakshy, Rosenn, Marlow, and
  Adamic}]{DBLP:journals/corr/abs-1201-4145}
Eytan Bakshy, Itamar Rosenn, Cameron Marlow, and Lada Adamic. 2012.
\newblock \href {https://doi.org/10.1145/2187836.2187907} {The role of social
  networks in information diffusion}.
\newblock In \emph{Proceedings of the 21st International Conference on World
  Wide Web}, WWW ’12, page 519–528, New York, NY, USA. Association for
  Computing Machinery.

\bibitem[{Bamman et~al.(2020)Bamman, Lewke, and Mansoor}]{bamman2019annotated}
David Bamman, Olivia Lewke, and Anya Mansoor. 2020.
\newblock \href {https://www.aclweb.org/anthology/2020.lrec-1.6} {An annotated
  dataset of coreference in {E}nglish literature}.
\newblock In \emph{Proceedings of The 12th Language Resources and Evaluation
  Conference}, pages 44--54, Marseille, France. European Language Resources
  Association.

\bibitem[{Bamman et~al.(2019)Bamman, Popat, and Shen}]{literaryentities}
David Bamman, Sejal Popat, and Sheng Shen. 2019.
\newblock \href {https://doi.org/10.18653/v1/N19-1220} {An annotated dataset of
  literary entities}.
\newblock In \emph{Proceedings of the 2019 Conference of the North {A}merican
  Chapter of the Association for Computational Linguistics: Human Language
  Technologies, Volume 1 (Long and Short Papers)}, pages 2138--2144,
  Minneapolis, Minnesota. Association for Computational Linguistics.

\bibitem[{Bamman et~al.(2014)Bamman, Underwood, and Smith}]{bammanliterary2014}
David Bamman, Ted Underwood, and Noah~A. Smith. 2014.
\newblock A {B}ayesian mixed effects model of literary character.
\newblock In \emph{Proceedings of the 52nd Annual Meeting of the Association
  for Computational Linguistics (Volume 1: Long Papers)}, pages 370--379,
  Baltimore, Maryland. Association for Computational Linguistics.

\bibitem[{Berg-Kirkpatrick et~al.(2012)Berg-Kirkpatrick, Burkett, and
  Klein}]{berg2012empirical}
Taylor Berg-Kirkpatrick, David Burkett, and Dan Klein. 2012.
\newblock An empirical investigation of statistical significance in {NLP}.
\newblock In \emph{Proceedings of the 2012 Joint Conference on Empirical
  Methods in Natural Language Processing and Computational Natural Language
  Learning}, pages 995--1005. Association for Computational Linguistics.

\bibitem[{Bourdieu(1996)}]{bourdieu1996rules}
P.~Bourdieu. 1996.
\newblock \emph{The Rules of Art: Genesis and Structure of the Literary Field}.
\newblock Meridian (Stanford, Calif.). Stanford University Press.

\bibitem[{Brendel et~al.(2011)Brendel, Meibauer, and Steinbach}]{quote2011}
Elke Brendel, J\"org Meibauer, and Markus Steinbach, editors. 2011.
\newblock \emph{Understanding Quotation}.
\newblock De Gruyter.

\bibitem[{Chaturvedi et~al.(2017)Chaturvedi, Iyyer, and {Daum\'{e}
  III}}]{Chaturvedi:Iyyer:Daume-III-2016}
Snigdha Chaturvedi, Mohit Iyyer, and Hal {Daum\'{e} III}. 2017.
\newblock Unsupervised learning of evolving relationships between literary
  characters.
\newblock In \emph{Association for the Advancement of Artificial Intelligence}.

\bibitem[{Coll~Ardanuy and
  Sporleder(2014)}]{coll-ardanuy-sporleder-2014-structure}
Mariona Coll~Ardanuy and Caroline Sporleder. 2014.
\newblock \href {https://doi.org/10.3115/v1/W14-0905} {Structure-based
  clustering of novels}.
\newblock In \emph{Proceedings of the 3rd Workshop on Computational Linguistics
  for Literature ({CLFL})}, pages 31--39, Gothenburg, Sweden. Association for
  Computational Linguistics.

\bibitem[{Del~Vicario et~al.(2016)Del~Vicario, Bessi, Zollo, Petroni, Scala,
  Caldarelli, Stanley, and Quattrociocchi}]{del2016spreading}
Michela Del~Vicario, Alessandro Bessi, Fabiana Zollo, Fabio Petroni, Antonio
  Scala, Guido Caldarelli, H~Eugene Stanley, and Walter Quattrociocchi. 2016.
\newblock The spreading of misinformation online.
\newblock \emph{Proceedings of the National Academy of Sciences},
  113(3):554--559.

\bibitem[{Devlin et~al.(2019)Devlin, Chang, Lee, and
  Toutanova}]{devlin-etal-2019-bert}
Jacob Devlin, Ming-Wei Chang, Kenton Lee, and Kristina Toutanova. 2019.
\newblock {BERT}: Pre-training of deep bidirectional transformers for language
  understanding.
\newblock In \emph{Proceedings of the 2019 Conference of the North {A}merican
  Chapter of the Association for Computational Linguistics: Human Language
  Technologies, Volume 1 (Long and Short Papers)}, pages 4171--4186,
  Minneapolis, Minnesota. Association for Computational Linguistics.

\bibitem[{Dror et~al.(2018)Dror, Baumer, Shlomov, and
  Reichart}]{dror-etal-2018-hitchhikers}
Rotem Dror, Gili Baumer, Segev Shlomov, and Roi Reichart. 2018.
\newblock The hitchhiker{'}s guide to testing statistical significance in
  natural language processing.
\newblock In \emph{Proceedings of the 56th Annual Meeting of the Association
  for Computational Linguistics (Volume 1: Long Papers)}, pages 1383--1392,
  Melbourne, Australia. Association for Computational Linguistics.

\bibitem[{Efron(1982)}]{efron1982jackknife}
Bradley Efron. 1982.
\newblock \emph{The jackknife, the bootstrap, and other resampling plans},
  volume~38.
\newblock Siam.

\bibitem[{Elson et~al.(2010)Elson, Dames, and
  McKeown}]{elson:2010:esn:1858681.1858696}
David~K. Elson, Nicholas Dames, and Kathleen~R. McKeown. 2010.
\newblock \href {http://dl.acm.org/citation.cfm?id=1858681.1858696} {Extracting
  social networks from literary fiction}.
\newblock In \emph{Proceedings of the 48th Annual Meeting of the Association
  for Computational Linguistics}, ACL '10, pages 138--147, Stroudsburg, PA,
  USA. Association for Computational Linguistics.

\bibitem[{Elson and McKeown(2010)}]{Elson:2010:AAQ:2898607.2898769}
David~K. Elson and Kathleen~R. McKeown. 2010.
\newblock Automatic attribution of quoted speech in literary narrative.
\newblock In \emph{Proceedings of the Twenty-Fourth AAAI Conference on
  Artificial Intelligence}, AAAI'10, pages 1013--1019. AAAI Press.

\bibitem[{Foster(2004)}]{foster2004}
Eric~K. Foster. 2004.
\newblock Research on gossip: Taxonomy, methods, and future directions.
\newblock \emph{Review of General Psychology}, 8(2).

\bibitem[{Friggeri et~al.(2014)Friggeri, Adamic, Eckles, and
  Cheng}]{friggeri2014rumor}
Adrien Friggeri, Lada Adamic, Dean Eckles, and Justin Cheng. 2014.
\newblock Rumor cascades.
\newblock In \emph{Eighth International AAAI Conference on Weblogs and Social
  Media}.

\bibitem[{Granovetter(1973)}]{granovetter1973strength}
Mark~S Granovetter. 1973.
\newblock The strength of weak ties.
\newblock \emph{American Journal of Sociology}, 78(6):1360--1380.

\bibitem[{Gruhl et~al.(2004)Gruhl, Guha, Liben-Nowell, and
  Tomkins}]{gruhl2004information}
Daniel Gruhl, Ramanathan Guha, David Liben-Nowell, and Andrew Tomkins. 2004.
\newblock Information diffusion through blogspace.
\newblock In \emph{Proceedings of the 13th international conference on World
  Wide Web}, pages 491--501. ACM.

\bibitem[{He et~al.(2013)He, Barbosa, and
  Kondrak}]{he-barbosa-kondrak:2013:ACL2013}
Hua He, Denilson Barbosa, and Grzegorz Kondrak. 2013.
\newblock Identification of speakers in novels.
\newblock In \emph{Proceedings of the 51st Annual Meeting of the Association
  for Computational Linguistics (Volume 1: Long Papers)}, pages 1312--1320,
  Sofia, Bulgaria. Association for Computational Linguistics.

\bibitem[{Hovy et~al.(2006)Hovy, Marcus, Palmer, Ramshaw, and
  Weischedel}]{hovy_ontonotes:_2006}
Eduard Hovy, Mitchell Marcus, Martha Palmer, Lance Ramshaw, and Ralph
  Weischedel. 2006.
\newblock {OntoNotes:} the 90\% solution.
\newblock In \emph{Proceedings of the Human Language Technology Conference of
  the {NAACL}, Companion Volume: Short Papers}, {NAACL-Short} '06, pages
  57--60, Stroudsburg, {PA}, {USA}. Association for Computational Linguistics.

\bibitem[{Iyyer et~al.(2016)Iyyer, Guha, Chaturvedi, Boyd-Graber, and
  {Daum\'{e} III}}]{Iyyer:Guha:Chaturvedi:Boyd-Graber:Daume-III-2016}
Mohit Iyyer, Anupam Guha, Snigdha Chaturvedi, Jordan Boyd-Graber, and Hal
  {Daum\'{e} III}. 2016.
\newblock Feuding families and former friends: Unsupervised learning for
  dynamic fictional relationships.
\newblock In \emph{North American Association for Computational Linguistics}.

\bibitem[{Jannidis et~al.(2016)Jannidis, Reger, Krug, Weimer, Macharowsky, and
  Puppe}]{jannidis2016}
F.~Jannidis, I.~Reger, M.~Krug, L.~Weimer, L.~Macharowsky, and F.~Puppe. 2016.
\newblock Comparison of methods for the identification of main characters in
  {G}erman novels.
\newblock In \emph{Digital Humanities}.

\bibitem[{Kwon et~al.(2013)Kwon, Cha, Jung, Chen, and Wang}]{kwon2013prominent}
Sejeong Kwon, Meeyoung Cha, Kyomin Jung, Wei Chen, and Yajun Wang. 2013.
\newblock Prominent features of rumor propagation in online social media.
\newblock In \emph{2013 IEEE 13th International Conference on Data Mining},
  pages 1103--1108. IEEE.

\bibitem[{Labatut and Bost(2019)}]{labatut}
Vincent Labatut and Xavier Bost. 2019.
\newblock \href {http://arxiv.org/abs/1907.02704} {Extraction and analysis of
  fictional character networks: {A} survey}.
\newblock \emph{CoRR}, abs/1907.02704.

\bibitem[{Lee and Yeung(2012)}]{lee-yeung-2012-extracting}
John Lee and Chak~Yan Yeung. 2012.
\newblock Extracting networks of people and places from literary texts.
\newblock In \emph{Proceedings of the 26th Pacific Asia Conference on Language,
  Information, and Computation}, pages 209--218, Bali, Indonesia. Faculty of
  Computer Science, Universitas Indonesia.

\bibitem[{Lee et~al.(2017)Lee, He, Lewis, and Zettlemoyer}]{lee-etal-2017-end}
Kenton Lee, Luheng He, Mike Lewis, and Luke Zettlemoyer. 2017.
\newblock End-to-end neural coreference resolution.
\newblock In \emph{Proceedings of the 2017 Conference on Empirical Methods in
  Natural Language Processing}, pages 188--197, Copenhagen, Denmark.
  Association for Computational Linguistics.

\bibitem[{Leskovec et~al.(2009)Leskovec, Backstrom, and
  Kleinberg}]{leskovec2009meme}
Jure Leskovec, Lars Backstrom, and Jon Kleinberg. 2009.
\newblock Meme-tracking and the dynamics of the news cycle.
\newblock In \emph{Proceedings of the 15th ACM SIGKDD international conference
  on Knowledge discovery and data mining}, pages 497--506. ACM.

\bibitem[{Leskovec et~al.(2007)Leskovec, McGlohon, Faloutsos, Glance, and
  Hurst}]{leskovec2007patterns}
Jure Leskovec, Mary McGlohon, Christos Faloutsos, Natalie Glance, and Matthew
  Hurst. 2007.
\newblock Patterns of cascading behavior in large blog graphs.
\newblock In \emph{Proceedings of the 2007 SIAM international conference on
  data mining}, pages 551--556. SIAM.

\bibitem[{Levine(2009)}]{levine2009}
Caroline Levine. 2009.
\newblock Narrative networks: Bleak house and the affordances of form.
\newblock \emph{Novel: A Forum on Fiction}, 42(3).

\bibitem[{Makazhanov et~al.(2014)Makazhanov, Barbosa, and
  Kondrak}]{DBLP:journals/corr/MakazhanovBK14}
Aibek Makazhanov, Denilson Barbosa, and Grzegorz Kondrak. 2014.
\newblock Extracting family relationship networks from novels.
\newblock \emph{CoRR}, abs/1405.0603.

\bibitem[{Margolis(2012)}]{margolis2012}
Stacey Margolis. 2012.
\newblock Network theory circa 1800: {Charles Brockden Brown}'s ``{Arthur
  Mervyn}''.
\newblock \emph{Novel: A Forum on Fiction}, 45(3).

\bibitem[{Martin(2014)}]{martin2014}
Nicholas Martin. 2014.
\newblock Literature and gossip: An introduction.
\newblock \emph{Forum for Modern Language Studies}, 50(2).

\bibitem[{Mazanec(2018)}]{mazanec2018networks}
Thomas~J Mazanec. 2018.
\newblock Networks of exchange poetry in late medieval {C}hina: Notes toward a
  dynamic history of {T}ang literature.
\newblock \emph{Journal of Chinese Literature and Culture}, 5(2):322--359.

\bibitem[{Miller and Hagberg(2011)}]{10.1007/978-3-642-21286-4_10}
Joel~C. Miller and Aric Hagberg. 2011.
\newblock Efficient generation of networks with given expected degrees.
\newblock In \emph{Algorithms and Models for the Web Graph}, pages 115--126,
  Berlin, Heidelberg. Springer Berlin Heidelberg.

\bibitem[{Moretti(2011)}]{moretti2011network}
Franco Moretti. 2011.
\newblock \emph{Network theory, plot analysis}.
\newblock Stanford Literary Lab.

\bibitem[{Muzny et~al.(2017)Muzny, Fang, Chang, and Jurafsky}]{muzny2017two}
Grace Muzny, Michael Fang, Angel Chang, and Dan Jurafsky. 2017.
\newblock A two-stage sieve approach for quote attribution.
\newblock In \emph{Proceedings of the 15th Conference of the European Chapter
  of the Association for Computational Linguistics: Volume 1, Long Papers},
  volume~1, pages 460--470.

\bibitem[{Niculae et~al.(2015)Niculae, Suen, Zhang, Danescu-Niculescu-Mizil,
  and Leskovec}]{niculae2015quotus}
Vlad Niculae, Caroline Suen, Justine Zhang, Cristian Danescu-Niculescu-Mizil,
  and Jure Leskovec. 2015.
\newblock Quotus: The structure of political media coverage as revealed by
  quoting patterns.
\newblock In \emph{WWW}.

\bibitem[{Nivre et~al.(2016)Nivre, de~Marneffe, Ginter, Goldberg, Hajic,
  Manning, McDonald, Petrov, Pyysalo, Silveira et~al.}]{nivre2016universal}
Joakim Nivre, Marie-Catherine de~Marneffe, Filip Ginter, Yoav Goldberg, Jan
  Hajic, Christopher~D Manning, Ryan McDonald, Slav Petrov, Sampo Pyysalo,
  Natalia Silveira, et~al. 2016.
\newblock Universal dependencies v1: A multilingual treebank collection.
\newblock In \emph{Proceedings of the 10th International Conference on Language
  Resources and Evaluation (LREC 2016)}, pages 1659--1666.

\bibitem[{Piper et~al.(2017)Piper, Algee-Hewitt, Sinha, Ruths, and
  Vala}]{piper2017}
Andrew Piper, Mark Algee-Hewitt, Koustuv Sinha, Derek Ruths, and Hardik Vala.
  2017.
\newblock Studying literary characters and character networks.
\newblock In \emph{Digital Humanities}.

\bibitem[{Schantz(2008)}]{schantz2008}
Ned Schantz. 2008.
\newblock \emph{Gossip, Letters, Phones: The Scandal of Female Networks in Film
  and Literature}.
\newblock Oxford University Press.

\bibitem[{Sedgwick(1985)}]{Sedgwick1985}
Eve Sedgwick. 1985.
\newblock \emph{Between Men: English Literature and Male Homosocial Desire}.
\newblock Columbia University Press.

\bibitem[{Selisker(2015)}]{Selisker2015}
Scott Selisker. 2015.
\newblock The bechdel test and the social form of character networks.
\newblock \emph{New Literary History}, 46(3).

\bibitem[{So and Long(2013)}]{so2013network}
Richard~Jean So and Hoyt Long. 2013.
\newblock Network analysis and the sociology of modernism.
\newblock \emph{boundary 2}, 40(2):147--182.

\bibitem[{Spacks(1982)}]{spacks82}
Patricia Spacks. 1982.
\newblock In praise of gossip.
\newblock \emph{The Hudson Review}, 35(1).

\bibitem[{Spacks(1985)}]{spacks85}
Patricia Spacks. 1985.
\newblock \emph{Gossip}.
\newblock Knopf.

\bibitem[{Stenetorp et~al.(2012)Stenetorp, Pyysalo, Topi\'{c}, Ohta, Ananiadou,
  and Tsujii}]{Stenetorp:2012:BWT:2380921.2380942}
Pontus Stenetorp, Sampo Pyysalo, Goran Topi\'{c}, Tomoko Ohta, Sophia
  Ananiadou, and Jun'ichi Tsujii. 2012.
\newblock {BRAT}: A web-based tool for {NLP}-assisted text annotation.
\newblock In \emph{Proceedings of the Demonstrations at the 13th Conference of
  the European Chapter of the Association for Computational Linguistics}, EACL
  '12, pages 102--107, Stroudsburg, PA, USA. Association for Computational
  Linguistics.

\bibitem[{Sudhahar and Cristianini(2013)}]{Sudhahar2013}
Saatviga Sudhahar and Nello Cristianini. 2013.
\newblock Automated analysis of narrative content for digital humanities.
\newblock In \emph{International Journal of Advanced Computer Science}.

\bibitem[{Vosoughi et~al.(2018)Vosoughi, Roy, and Aral}]{vosoughi2018spread}
Soroush Vosoughi, Deb Roy, and Sinan Aral. 2018.
\newblock The spread of true and false news online.
\newblock \emph{Science}, 359(6380):1146--1151.

\bibitem[{Wilkerson et~al.(2015)Wilkerson, Smith, and
  Stramp}]{wilkerson2015tracing}
John Wilkerson, David Smith, and Nicholas Stramp. 2015.
\newblock Tracing the flow of policy ideas in legislatures: A text reuse
  approach.
\newblock \emph{American Journal of Political Science}, 59(4):943--956.

\bibitem[{Woolf(1929)}]{woolf1929}
Virginia Woolf. 1929.
\newblock \emph{A Room of One's Own}.
\newblock Harcourt Brace Jovanovich.

\end{thebibliography}

\clearpage

\end{document}